# Benchmarking Robustness of Deep Learning Classifiers Using Two-Factor Perturbation

Wei Dai
Department of Computer Science
Southeast Missouri State University
Cape Girardeau, MO, USA
wdai@semo.edu

Daniel Berleant
Department of Information Science
University of Arkansas at Little Rock
Little Rock, AR, USA
jdberleant@ualr.edu

*Abstract*— **Accuracies of deep learning (DL) classifiers are often unstable in that they may change significantly when retested on adversarial images, imperfect images, or perturbed images. This paper adds to the fundamental body of work on benchmarking the robustness of DL classifiers on defective images. To measure robust DL classifiers, previous research reported on single-factor corruption. We created comprehensive 69 benchmarking image sets, including a clean set, sets with single factor perturbations, and sets with two-factor perturbation conditions. The state-of-the-art two-factor perturbation includes (a) two digital perturbations (salt & pepper noise and Gaussian noise) applied in both sequences, and (b) one digital perturbation (salt & pepper noise) and a geometric perturbation (rotation) applied in both sequences. Previous research evaluating DL classifiers has often used top-1/top-5 accuracy. We innovate a new two-dimensional, statistical matrix to evaluating robustness of DL classifiers. Also, we introduce a new visualization tool, including minimum accuracy, maximum accuracy, mean accuracies, and coefficient of variation (CV), for benchmarking robustness of DL classifiers. Comparing with single factor corruption, we first report that using two-factor perturbed images improves robustness of DL classifiers.**

## I. INTRODUCTION

Computer scientists and engineers have created many benchmarking tools to measure hardware devices and compare software algorithms. Well known, non-profit benchmark organizations include Standard Performance Evaluation Corporation (SPEC) [1], Transaction Processing Performance Council (TPC) [2], Storage Performance Council (SPC) [3], and Machine Learning Performance (MLPerf) [4]. These evaluate CPUs, databases, storage, and machine learning respectively. MLPerf has measured the training and inference performance of ML hardware, software, and services. All these organizations update their benchmark tools, retire obsolete benchmark programs, and publish documentation on their websites.

In the ML field, deep learning (DL) classifiers have been shown to work well on sets of high quality images. Deep learning (DL) algorithms have dramatically expanded in use in many scientific fields and industrial projects [5]-[6]. For example, researchers discussed how to improve data quality via machine learning in recently reported results [7]-[8]. DL classifiers have classified objects in sets of high quality images at accuracies as high as 97.3%, which is better than human capabilities [9]. However, sets of *low* quality images present a more challenging, yet critically important, problem. If the data is imperfect, as real data so often are, how will results be affected? In [10], the researchers discovered that human visual systems are more robust than DL classifiers when images are manipulated by contrast reduction, additive noise, and distortions. In [11], DL classifier performance was lower than that of humans on recognizing images corrupted with Gaussian noise or Gaussian blur. Such limitations are a problem because, for example, imperfect sensors sending degraded images to DL applications could result in unexpected accidents. Thus, enabling DL classifiers to robustly handle imperfect images would have valuable benefits in such safety-critical application areas as unmanned vehicles, health care and life support systems [12]-[13].

DL classifiers can also make mistakes on apparently high quality images in the DL security field. For example, an adversarial image may be a high quality image modified by tiny perturbations chosen to confuse DL classifiers. Even though humans may not notice these perturbations, these deliberately modified images can confuse DL classifiers, reducing their accuracy [10][16][14]. Thus, the robustness of DL classifiers across image perturbation conditions will continue to merit attention. To measure robustness of DL classifiers, we can expect a robust DL classifier to have both high accuracy as well as low variance across the perturbation conditions.

In this paper, our contributions are as follows:

- Compared with using only clean images or single factor corrupted images, training with two-factor perturbation image sets can improve the robustness of DL classifiers better than existing approaches. To be best of our knowledge, this is the first report that two-factor perturbation improves robustness of DL classifiers. For more details, see Table III at the end.
- We provide a new visualization tool, named the mCV plot, for comparing DL classifiers. This complements standard tools like tables, line diagrams, and bar charts. With this tool we integrated mean accuracy and CV into a two-dimensional statistical visualization for quantitatively benchmarking robustness of DL classifiers. The mCV plot readily permits expressing such relevant traits of DL classifiers as minimal/maximal accuracy, accuracy, mean accuracies, and CV. For details, see Eqs. 1-4.
- Unlike previous single-factor perturbation work, we provide state-of-the-art two-factor perturbations that include two digital perturbations (salt & pepper noise and Gaussian noise) applied in both orders. Also, a two-factor perturbation may consist of one digital perturbation (salt & pepper noise) and a geometric perturbation (rotation) applied in both orders. Briefly, two-factor perturbation provides two natural perturbations on images. For some examples, see Figs. 1 and 2. We created test sets with 69 test images, one clean image and 68 corrupted images. Most perturbations involved corrupting images twice, by two different perturbations. These image sets and source codes are shared on a website (cslinux.semo.edu/david/data.html) to support additional research projects. To the best of our knowledge, this is a new approach to measuring robustness of deep learning classifiers.

The rest of this paper is structured as follows. Section II is Related Work. In Section III, we provide the research design and methodology. Section IV introduces the experimental results. Discussion, Future Work, and Conclusion constitute Section V, Section VI, and Section VII, respectively.

I. RELATED WORK

Researchers have discovered that DL classifiers are often fragile when faced with image perturbations. In [17], the authors demonstrated that the Google Cloud Vision API could be misled after adding approximately 14.25% impulse noise density. In [18], DL classifiers had reduced accuracy of classification when testing five types of quality distortions, including JPEG/JPEG2000 compression, blur, Gaussian noise, and contrast. There have been various previously reported corruptions imposed on high quality ImageNet datasets. In [19] the authors created 15 types of single perturbation, including noise, blur, and brightness. In [11][20], the authors used Gaussian noise and Gaussian blur to change images. While these research projects created defective images via single factor corruptions, in the real world, image quality would often be impacted by more than one perturbation. For example, both bad weather (e.g. fog) and vibration could together impact cameras of self-driven cars, resulting in unexpected traffic issues. In this research, we design 69 benchmarking image sets, including a clean set, sets with single factor perturbations, and sets with two-factor perturbation conditions for evaluating robustness in DL classifiers.

In addition to corruptions that involve a loss of information, vulnerable DL classifiers can also make mistakes on adversarial corruptions in the DL security field. An adversarial picture is a high quality image modified by tiny deliberate corruptions chosen to confuse DL classifiers. Even though humans may not notice these corruptions, research shows that adversarial examples can mislead DL classifiers with deliberately modified images, leading to confused classifiers and reduced accuracy of DL classifiers [14]–[16].

Benchmarking metrics have many applications. Previous researchers have measured the robustness of DL classifiers via top-1 and top-5 precision analyses [21]. For example [19] chose the average rates of correct ("top-1") and almost correct ("top-5") classifications, the authors setting the top-1 error rate of AlexNet as the reference error rate. Existing results demonstrated how to utilize top-$n$ to evaluate robustness of DL. However, we provide a two-dimensional metric, consisting of mean accuracy and coefficient of variation (CV). This supports characterizing the stability of DL classifiers as having high mean accuracy along with small coefficient variance of accuracies across perturbation conditions.

The coefficient of variation is the ratio of the standard deviation to the mean. The coefficient of variation is used in different fields, including physics, medicine, chemistry, and engineering [22][23]. In agriculture, Francis and Kannenberg [24] used the mean yield and coefficient of variation to analyzing yield stability. The CV offers a way to help differentiate the variation from the mean because, compared to using variance, CV controls for the mean. So, when judging a DL classifier, this article benchmarks both the average accuracy, $\bar{x}$, and the coefficient of variation, CV, as they say different things. A high accuracy and small CV for a DL classifier is better than a low accuracy with a large CV because it is better on both dimensions.

## II. RESEARCH DESIGN AND METHODOLOGY

### A. Two-factor perturbations

Previous research projects usually used single-factor corruptions on images [19][25] for benchmarking DL classifiers. For example, in [19] the authors created imperfect images through single-factor perturbations. However, images are typically impacted by multiple factors in the real world. For example, images captured by drones could be impacted by the drones' vibrations and environments such as fog or strong light. As another example, traffic signs could be faded, dirty, and/or rotated 5 or 10 degrees.

This project used images made to be defective with two-factor perturbation. Two-factor perturbation refers to modifying images using two types of perturbation, one after the other. The two-factor perturbation may include the two digital perturbations of salt & pepper noise and Gaussian noise in different sequences. Another case of two-factor perturbation consists of a geometric perturbation (rotation) followed by a noise perturbation or noise followed by rotation.

For instance, SP-Gaussian perturbation means we firstly used salt & pepper noise for corrupting images, then Gaussian noise for corrupting more. On the other hand, Gaussian-SP perturbation denotes changing images through Gaussian perturbation first, then applying salt & pepper perturbation. Examples of SP-Gaussian perturbation and Gaussian-SP perturbation are shown in Fig. 1. Examples of SP-Rotation perturbation and Rotation-SP perturbation are shown in Fig. 2.

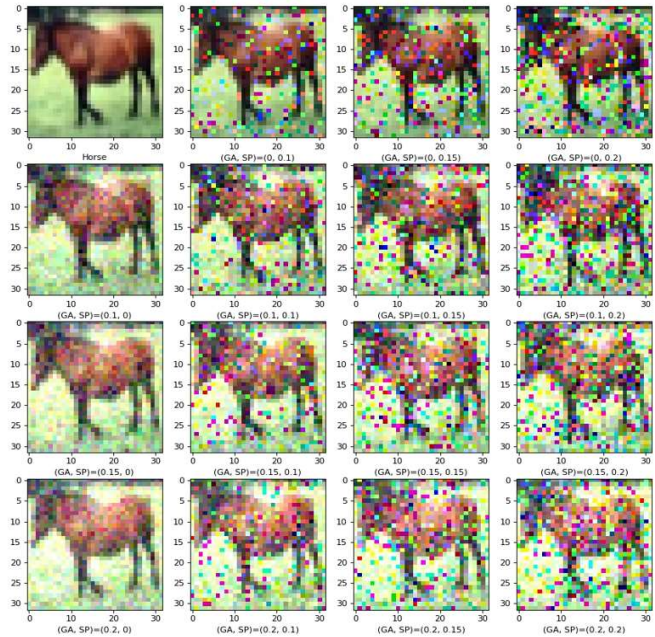

Fig. 1. Results of SP-Gaussian corruption. The original picture is from CIFAR-10 as shown on the top left corner. (SP, GA) denotes two-factor corruption with salt & pepper noise applied first, and then Gaussian noise. (SP, GA) = (0, 0.1) indicates salt & pepper noise density is zero, but Gaussian noise density is 0.10.

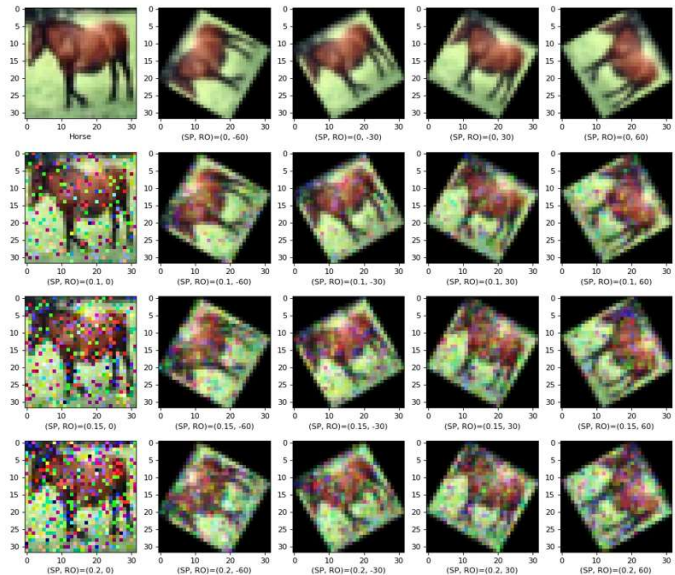

Fig. 2. Results of SP-Rotation corruption. The original picture locates at the top left corner. (SP, RO) means salt & pepper noise corruption was applied first, followed by rotation. (SP, RO) = (0.1, -30) indicates salt & pepper noise density is 0.1, and the image counterclockwise rotation was 30 degrees. (SP, RO) = (0.2, 60) indicates salt & pepper noise density is 0.2, and the image rotation is clockwise 60 degrees.

The two-factor perturbation occurs in two stages. We use perturbation level and rotation ranges of [0.1, 0.2] and [-60, +60] degrees, respectively. Zero degrees denotes that we do not rotate images, negative degrees means we rotate images counterclockwise, and positive degrees means we rotate images clockwise. Specifically, two-factor perturbation types SP-Gaussian, Gaussian-SP, SP-Rotation and Rotation-SP and their ranges will be applied as shown in Table I.

TABLE I. PERTURBATION SEQUENCES, PERTURBATION TYPES, AND NOISE VALUES

| Perturbed Sequence | Respective Noise Strengths | |
|---|---|---|
| SP-Gaussian | 0.1, 0.15 & 0.2 | 0.1, 0.15 & 0.2 |
| Gaussian-SP | 0.1, 0.15 & 0.2 | 0.1, 0.15 & 0.2 |
| SP-Rotation | 0.1, 0.15 & 0.2 | $-60^0, -30^0, 0^0, 30^0$ & $60$ |
| Rotation-SP | $-60^0, -30^0, 0^0, 30^0$ & $60^0$ | 0.1, 0.15 & 0.2 |

For measuring DL classifiers, we designed three steps: training, inference, and analysis. For details, see Fig. 3. In the training step, we used pretrained models [26] to train and fine-tune DL classifiers. We used the Image Processing Toolbox of MATLAB 2019b for the perturbed CIFAR-10 set. Both source codes and image sets are shared on the Internet. Readers may access resources at http://cslinux.semo.edu/david/data.html to support future academic research and industry projects.

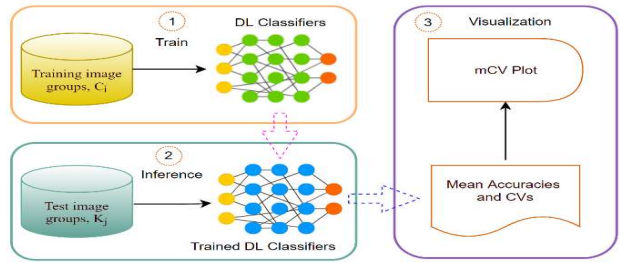

Fig. 3. An architecture for benchmarking DL classifiers. There are three steps for measuring the robustness of DL classifiers. Step 1 trains DL classifiers. Step 2 performs inference using the trained DL classifiers. Step 3 depicts the average accuracies and coefficients of variation (CV for short) with a mCV plot.

*B. Evaluating robust deep learning classifiers using the mCV plot*

To evaluate the robustness of DL classifiers, we start with the fundamental assumption that one DL classifier may be regarded as more robust than another if it has higher accuracy and smaller coefficient of variation. The coefficient of variation is derived from the standard deviation ($\sigma$), which is a measure of spread of distributions in statistics. A larger $\sigma$ denotes that the population of the sample data is further dispersed from the mean. A smaller $\sigma$ indicates the population of sample data is more clumped together.

In statistics, the formula for standard deviation ($\sigma$) is shown in Eq. 1.

$$\sigma = \sqrt{\frac{\sum_{i=1}^{n}(x_i - \mu)^2}{n}} \qquad (1)$$

where $x_i$, $n$, and $\mu$ are a sample value, the number of data and the mean, respectively.

When testing performance across a group of data sets, we can collect the mean of the accuracies, $\mu$, and the standard deviation of the accuracies, $\sigma$. The $\sigma$ of different DL classifiers should be compared when their $\mu$ values are similar, because if two DL classifiers have the same mean accuracies, a smaller variation ($\sigma$) for one shows better stability in its performance across different image quality conditions. We can say that a smaller variation ($\sigma$) indicates that the DL classifier performs more consistently. Ideally, we want a DL classifier to have high accuracy with small standard deviation.

The **c**oefficient of **v**ariation (or **CV**) is the standard deviation as a percentage of the mean as shown in Eq. 2.

$$CV\% = 100 \times \frac{\sigma}{\mu} \qquad (2)$$

where $\sigma$ is the population standard deviation, and the mean, $\mu$, is the average of the data sets.

Unlike one dimensional accuracy (e.g. top-1/top-5) measures of DL classifiers, we developed the mCV statistical visualization for comparing DL classifier robustness in two dimensions. This is a four-quadrant statistical plot approach which we call the mCV (for mean accuracy and coefficient of variation) plot. See Fig. 4. In this graphical plot, the $Y$-axis and $X$-axis indicate mean accuracy and coefficient of variation respectively. Then a reference point is chosen that splits the figure into four groups as shown in Fig. 4.

How to split DL classifiers into the four groups is given in Algorithm 1. Accu(clean) refers to the accuracy of a classifier when trained on clean images but tested on all of them. CV(clean) refers to the coefficient of variation under the same conditions. Experimental results are shown in Figs. 5-7. In Fig. 4, $AlexNet_{clean}^{alltests}$ is the reference point. $ResNet50_{clean}^{alltests}$ and $VGG19_{clean}^{alltests}$ are the reference points in Fig. 6 and Fig. 7, respectively.

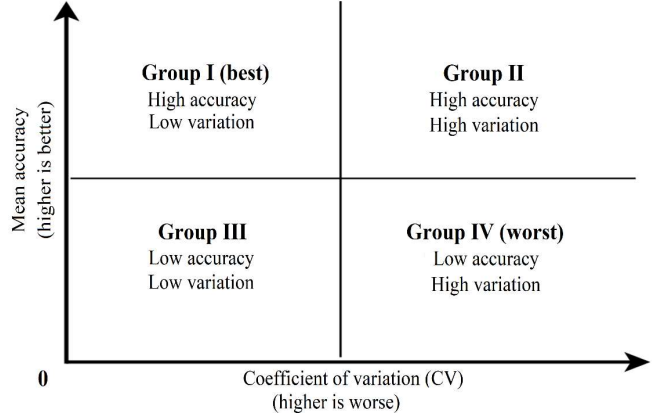

Fig. 4. Schema of mCV plots. The $Y$-axis and $X$-axis indicate mean accuracy and coefficient of variation. The mean accuracy and CV of the chosen reference point in the middle divides the above plot into four groups. The best group for a DL classifier to be in is Group I.

A trained DL classifier is notated as $D_{trainsets}^{testsets}$, where $D$ refers to the DL classifier's name, testsets refers to a group of test sets, and trainsets refers to a group of training sets. Then, we test the accuracy rate on every perturbation type $c$. Different $c$ values are different perturbation types, and level of severity is $s$, with $s \in \{0.1, 0.15, 0.2\}$ when doing salt and pepper (SP for short) or Gaussian (GA for short) perturbation. Also, $s \in \{-60°, -30°, 0°, 30°, 60°\}$ when doing rotation perturbation. The accuracy rate is written $Accu(D)$ as shown in Eq. 3.

$$Accu(D)\% = 100 \times \frac{\sum_{k=1}^{n} Accu(D_{trainset_c}^{testset_k})}{n-1},$$

where both *c* and *k* mean an image group, respectively. Each testing image group is listed on the online website. Values for *c* were type 1 (clean images), type 5 (SP0.1), type 6 (SP0.1GA0.1), type 24 (GA0.1), type 25 (GA0.1SP0.1), type 43 (SP0.1RL30), type 34 (SP0.1RR30), type 40 (RL30) and type 32 (RR30). SP0.1GA0.1 means SP perturbation was applied at the 0.1 level followed by GA perturbation at the 0.1 level, respectively.

---

**Algorithm 1: Place a DL classifier into a group in the mCV graph.**
Input 1: MA (deep learning **M**ean **A**ccuracy)
Input 2: CV (deep learning **C**oefficient of **V**ariation)
Output: group number
IdentifyGroup (MA, CV){
  rMA ← mean Accu(clean)
  rCV ← CV(clean)
  if (MA ≥ rMA and CV ≤ rCV) return "Group I"
  if (MA ≥ rMA and CV > rCV) return "Group II"
  if (MA < rMA and CV ≤ rCV) return "Group III"
  if (MA < rMA and CV > rCV) return "Group IV"
}

---

For example, we can define $AlexNet_{clean}^{alltests}$ to denote the AlexNet classifier trained on a clean image set, and tested on all 69 different test image sets. $AlexNet_{SP.1RL3}^{alltests}$ then denotes the AlexNet classifier trained on an image set corrupted by SP0.1 then corrupted more by RL30 (rotate left 30º).

We calculate coefficient of variation as

$$CV(D)\% = 100 \times \frac{\sigma}{\mu} \qquad (4)$$

$$= 100 \times \frac{\sqrt{\frac{\sum_{k=1}^{69}(Accu(D)_k - \overline{Accu(D)})^2}{n}}}{\overline{Accu(D)}}$$

where $CV(D)$ indicates the coefficient of variation exhibited by a DL classifier. $\overline{Accu(D)}$ denotes the mean accuracy of the DL classifier. $Accu(D)_k$ indicates the accuracy of DL classifier on test set *k*. Variable *k*, $k \in [1, 69]$, is the testing image group (for more information, see the online website).

The training data consisted of nine training sets, a clean image set and eight corrupted image sets. Each training set had 500 images. The clean image set contained the original images. The eight corrupted image sets were corrupted according to SP0.1, GA0.1, SP0.1GA0.1, GA0.1SP0.1, SP0.1RL30, SP0.1RR30, RL30 and RR30.

Each testing set contained one clean image group and 68 corrupted image groups. The clean image group contained original images. Each testing group had 500 images. The training sets and testing sets had no overlap. In other words, if an image existed in a training set, the image will not be found in any testing set.

Abbreviated notations might not be easily read in Figs. 5-7, so we wrote labels instead. For example, the label "AlexNet(clean)" denotes the AlexNet classifier trained on the unmodified CIFAR-10 image set, but the testing was on all 69 data sets, including the clean CIFAR-10 image set and 68 corrupted versions, which we state as $AlexNet_{clean}^{alltests}$. Table III shows results for different training data sets of CIFAR-10 on different rows.

As shown in Fig. 5, the mean accuracy of the AlexNet(SP0.1RL30) classifier is higher (better) than AlexNet(RL30), and the CV of the AlexNet(RL30) is higher (worse) than for AlexNet(SP0.1RL30). Similarly the minimum accuracy of AlexNet(SP0.1RL30) is higher than the minimum accuracy of AlexNet(RL30). Also, we found the same results when comparing the two maximum accuracies. However, the accuracy of AlexNet(SP0.1RL30) on uncorrupted images, 89.96% (red circle), is a bit smaller than the corresponding accuracy of AlexNet(RL30), 90.72%. For accuracy values, see Table III. The comparison is summarized visually in the mCV plot, which shows AlexNet(RL30) in Group II, while AlexNet(SP0.1RL30) is in Group I, the best quadrant because it has smaller CVs, higher mean accuracies, or both compared to the other quadrants. The mCV plot provides a visual comparison among different DL classifiers on the same test protocol.

## III. EXPERIMENTAL RESULTS

### A. Benchmarking AlexNet on corrupted images

AlexNet(clean) represents the AlexNet classifier trained on clean data. Fig. 5 shows the following.
- The training datasets affected prediction CVs of the AlexNet classifiers. After training AlexNet on corrupted images, the largest CVs (with the exception of AlexNet(RR30) and AlexNet(RL30)) were less than when training was limited to clean images. The lowest accuracies were also (with the same exceptions) greater than the lowest accuracy of AlexNet trained on clean images.
- Training AlexNet on corrupted images can improve the mean accuracy during testing. In particular, all Accu($AlexNet_{corrupt}^{alltests}$) values were higher than Accu($AlexNet_{clean}^{alltests}$). That is, tested accuracies tended to be better when training sets contain images that have been corrupted by degradation with common forms of noise and distortion.
- When testing AlexNet on clean images after training on corrupted images, accuracies tended to be lower than when training AlexNet on clean images. This is intuitively unsurprising, although the difference is relatively slight.

In brief, training AlexNet on corrupted images often improved robustness of AlexNet.

### B. Benchmarking ResNet50 on corrupted images

Fig. 6 indicates the following.
- The training datasets affected prediction CVs of the ResNet50 classifiers. After training ResNet50 classifiers on corrupted images, the CVs tended to be lower (better) than after training on clean images.
- The minimum, mean and maximum accuracies, tested after training on corrupted images were lower (worse) than after training on clean images.
- As might be expected, training ResNet50 on corrupted images led to lower accuracy when tested on uncorrupted images compared to training on uncorrupted, clean images.

Summing up, training ResNet50 on corrupted images tended to improve coefficient of variation (that is, stability, in the sense of consistency of performance) of ResNet50 but reduce accuracy.

### C. Benchmarking $VGG-19$ on corrupted images

Fig. 7 shows the following.
- CV($VGG-19_{clean}^{alltests}$) is greater than any CV($VGG-19_{corrupt}^{alltests}$). This illustrates the positive effect of training on corrupted images to boost DL robustness.
- Only mean Accu($VGG-19_{GA0.1SP0.1}^{alltests}$) is slightly higher than mean Accu($VGG-19_{clean}^{alltests}$), while the others are lower.
- When tested on uncorrupted images, training on uncorrupted images works better than training on corrupted images.

Thus, training $VGG-19$ on corrupted images tends to reduce accuracy on uncorrupted images but improve coefficient of variation.

## IV. DISCUSSION

Our tests demonstrated that high accuracy of DL classifiers on high quality image sets does not ensure high stability of DL classifiers, as measured by coefficient of variation, when the classifiers were then tested on images of lowered quality, as is commonly the case in real world situations. The merged performance data of three DL classifiers with 27 training sets is shown in Table III. We conclude as follows.

- Most CV($D_{two-factor\ corrupt}^{alltests}$) values are less than the corresponding CV($D_{single\ factor\ corrupt}^{alltests}$) values. According to Table III, mean CV($D_{clean}^{alltests}$) is 2.94%, mean CV($D_{single-factor\ corrupt}^{alltests}$) is 1.82%, and mean CV($D_{two-factor\ corrupt}^{alltests}$) is 1.42%. Thus, mean CV($D_{two-factor\ corrupt}^{alltests}$) is somewhat better than mean CV($D_{single-factor\ corrupt}^{alltests}$), reducing it by 28.87%. This indicates that two-factor perturbation is better than single factor corruption for reducing coefficient of variation.
- Mean Accu($D_{clean}^{alltests}$) is 88.31%, mean Accu($D_{single-facto\ corrupt}^{alltests}$) is 87.10%, and mean Accu($D_{two-factor\ corrupt}^{alltests}$) is 87.36%. This suggests that training on clean, single-factor corrupted, and two-factor corrupted images do not differ greatly in later accuracy during testing.
- Minimal Accu($D_{clean}^{alltests}$) is 85.01%, minimal Accu($D_{single-facto\ corrupt}^{alltests}$) is 84.76%, and minimal Accu($D_{two-factor\ corrupt}^{alltests}$) is 89.57%. This indicates that minimal Accu($D_{two-factor\ corrupt}^{alltests}$) was better than Accu($D_{single-factor\ corrupt}^{alltests}$), increasing it by 5.37%. Briefly, two-factor perturbation was the best on this question. This result is consistent with the increased stability (lower CV) shown by training on corrupted images.
- Maximal Accu($D_{clean}^{alltests}$) is 92.83%, maximal Accu($D_{single-facto\ corrupt}^{alltests}$) is 90.01%, and maximal Accu($D_{two-factor\ corrupt}^{alltests}$) is 89.57%. To sum up, two-factor perturbation was slightly worse than single factor corruption, a result which is consistent with the increased stability of training on corrupted images.
- For different DL classifiers, the same corrupted image training sets may reduce or improve the mean accuracy of DL classifiers, depending on the classifier.
- Comparing clean and corrupted training image sets, Accu($D_{clean}^{clean}$), unsurprisingly, for most types of corruption is above Accu($D_{corrupt}^{clean}$).

To better understand the correlation coefficient, mean accuracy, and Accu(clean) which is the accuracy when trained on clean data and tested on all data, the Spearman's rank correlation coefficient and Pearson correlation coefficient were used. See Table II. Spearman's rank correlation coefficient is a statistic used for calculating the strength of a monotonic relationship between paired data (*X*, *Y*) in a sample *N*. Assume that a value of *X* has rank *K* and its corresponding value of *Y* has rank *L*. Then Spearman's rank $r_{xy}$ can be calculated for the sample data *N* as shown in Eq. 5.

$$r_{xy} = 1 - \frac{6\sum d_{xy}^2}{n(n^2-1)} \quad (5)$$

where *d* is the difference between *K* and *L*, $d_{xy}^2 = (K - L)^2$, and the number of data in *N* is *n*.

The Pearson correlation coefficient is used for evaluating the strength of a linear relationship between two random variables. The Pearson correlation coefficient, $r_{xy}$, may be calculated as shown in Eq. 6.

$$r_{xy} = \frac{\sum_i^n (x_i - \bar{x})(y_i - \bar{y})}{\sqrt{\sum_i^n (x_i - \bar{x})^2} \sqrt{\sum_i^n (y_i - \bar{y})^2}} \quad (6)$$

where $x_i, y_i$ are paired datasets indexed with *i*, and $\bar{x}, \bar{y}$ are means. The sample size is *n*.

Table II shows that the Pearson correlation coefficients are all below 0.5. This is relatively low, indicating that CV, mean accuracy, and Accu(clean) are not strongly dependent on each other.

A robust DL classifier [19] [21] should produce results with high accuracies and small CVs. While accuracy and CV are each one-dimensional metrics, we integrated them both into a two-dimensional metric by designing and using the mCV plot for visualizing DL classifier robustness. An advantage of the mCV plot is that it provides a readily understandable visualization tool for measuring DL classifier robustness. This complements standard tools used in related work like tables, line diagrams, and bar charts. The mCV plot visualization makes the results of the present study, the first to focus on using two-factor perturbations to images for benchmarking DL classifiers, easy to compare in a convenient, understandable, summarized format.

TABLE II. SPEARMAN'S RANK & PEARSON CORRELATION COEFFICIENTS.

| Data Sets | Spearman's Rank | Pearson Correlation Coefficient |
|---|---|---|
| CV & mean Accu($D$) | 0.202 | 0.096 |
| CV & Accu(clean) | 0.345 | 0.365 |
| Mean accuracy & Accu(clean) | 0.057 | 0.052 |

This investigation required 69 benchmarking image sets, including a clean set, sets with single factor perturbations, and sets with two-factor perturbation conditions.

There are some potential limitations in this research. The first is the datasets. The CIFAR-10 image sets are relatively small. Also, the CIFAR-10 data does not have grayscale pictures. The second limitation concerns the 68 corrupted image sets that were corrupted by SP perturbations, GA perturbations, and rotations. There are other types of imperfections which were not tested.

In summary, previous researchers have indicated that corrupted image sets reduce the accuracy of classifiers [18] or lead to underfitting [21]. We have found that this appears to be a result of using only high quality image sets for training. As an alternative, our results indicate the advantage of using corrupted image sets during training. In addition, our results were partially consistent with [18] and [21]. When testing on clean images, our results were consistent with previous researchers' reports. Nevertheless, when choosing 69 testing image sets, our results confirmed that training on defective images degraded by two-factor perturbations can enhance DL classifier robustness.

## V. CONCLUSIONS AND FUTURE WORK

We tested three DL classifiers on images corrupted by two-factor perturbations. The research demonstrated that training on two-factor corrupted images can improve coefficient of variation in the performance of all three classifiers, in terms of consistency of performance across data sets. As a consequence, accuracy on the test set for which the classifiers performed worst was usually better for classifiers trained on corrupted images than for classifiers trained on clean images for AlexNet and VGG-19, even though for VGG-199 mean accuracy across data sets was usually lower when trained on corrupted images. ResNet accuracy did not benefit from training on corrupt images. Thus, imperfect images should certainly be considered in training, testing and benchmarking DL classifiers. In particular, any domain for which occasional poor performance should be avoided to the extent possible — even if average performance is relatively high — is a good candidate for application of the approach. Candidate domains include those where even occasional cases of poor performance can be costly, such as vision systems in self-driving vehicles, detecting serious diseases in medical images, etc.

In this research, we provided a two-dimensional visualization and metric that integrates minimum/maximum accuracy, mean accuracy, and coefficient of variation (CV). This is the mCV plot. It is useful for benchmarking robustness of DL classifiers. We investigated robustness of example classifiers based on training and testing using image sets that have been deliberately corrupted to model the imperfections of images inherent in many real world applications. Future classifiers could also be tested this way, as well as designed specifically to do well on robustness benchmarking. This study serves as a proof of concept that accuracies and CVs can be used together for quantitatively benchmarking robustness in the performance of DL classifiers.

This article explored deep learning algorithm robustness under two-factor corruption. However, the results lead naturally to other questions that will require further research before this important issue could be considered fully explored. These include the following.

1) *Data sets*. There are many and varied image sets on which this work might be applied and it is not known yet how well the results might generalize to data sets with different characteristics and from different domains. This methodological limitation remains an potential question in this type of research.

2) *Noise conditions*. This article explored two-factor perturbation with rotation, salt & pepper noise, and Gaussian noise on three data sets, but other types of perturbations occur in real image processing problems and may also impact DL classifier robustness. Also, different noise sequences may affect results.

3) *DL models*. We tested the AlexNet, ResNet50, and VGG-19 classifier algorithms. Algorithm development is a key feature of current work in the field and presents important questions related to the work we present. Certainly, a natural next step after algorithm benchmarking is development of new and better algorithms.

4) *DL theories*. We have empirically investigated two-factor image corruption, which leads naturally to a number of questions.
   - Why can training on imperfect images improve the coefficient of variation, and thus the consistency of performance, of DL classifiers?
   - How can we balance the accuracy and consistency of DL classifiers?
   - How can we design DL classifiers to achieve both higher accuracy and smaller CV? Such algorithmic developments may be expected in the future as the DL field rapidly progresses.

5) *Broadened impact*. We evaluated three DL models. Can the benchmarking approach and the mCV plot also be used to test audio data sets, numerical data, and regression and other statistical models?

6) *From 2-factor to N-factor corruption stacking*. We have noted that 2-factor corruption better models the real world of images with their diverse imperfections. Obviously real images are subject to various other corruptions as well. Is 2-factor image perturbation ideal for benchmarking, or would *N*-factor perturbation for some value(s) of $N \neq 2$ work better?

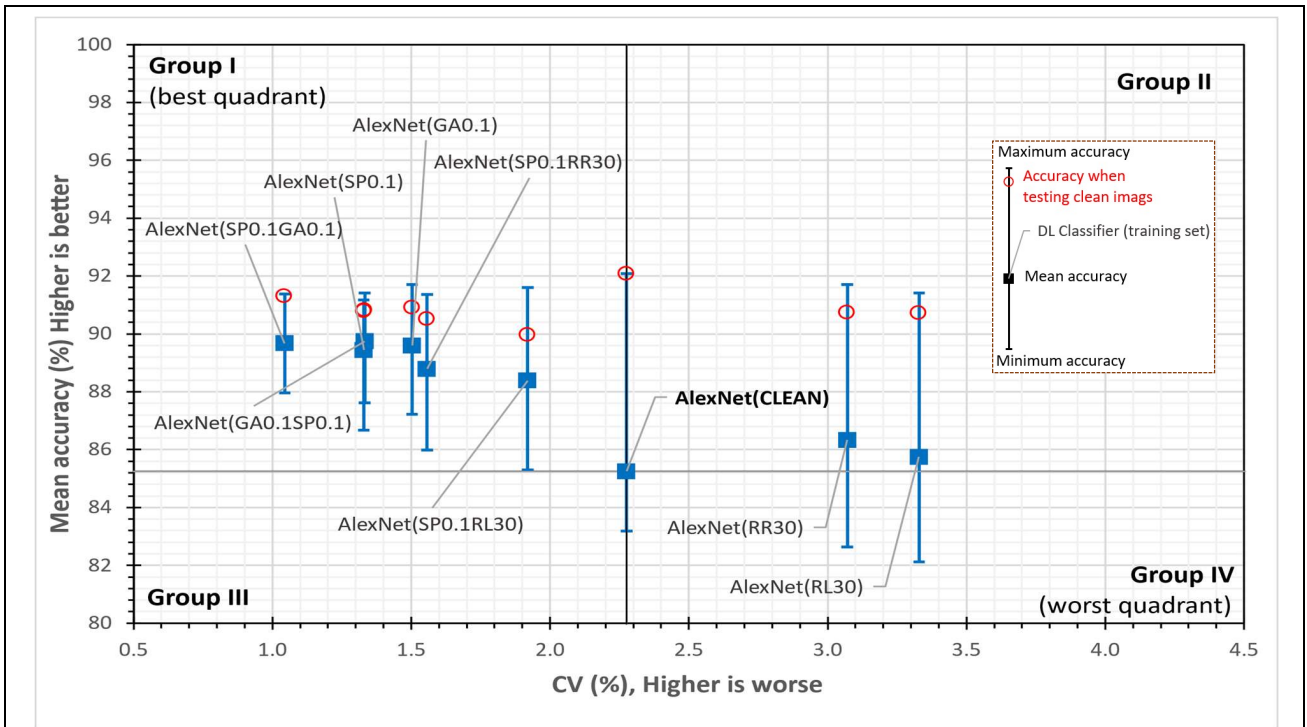

Fig. 5. The mCV plot for evaluating robustness of AlexNet classifiers after corrupted image groups were used to train the DL classifier on the CIFAR-10 dataset.

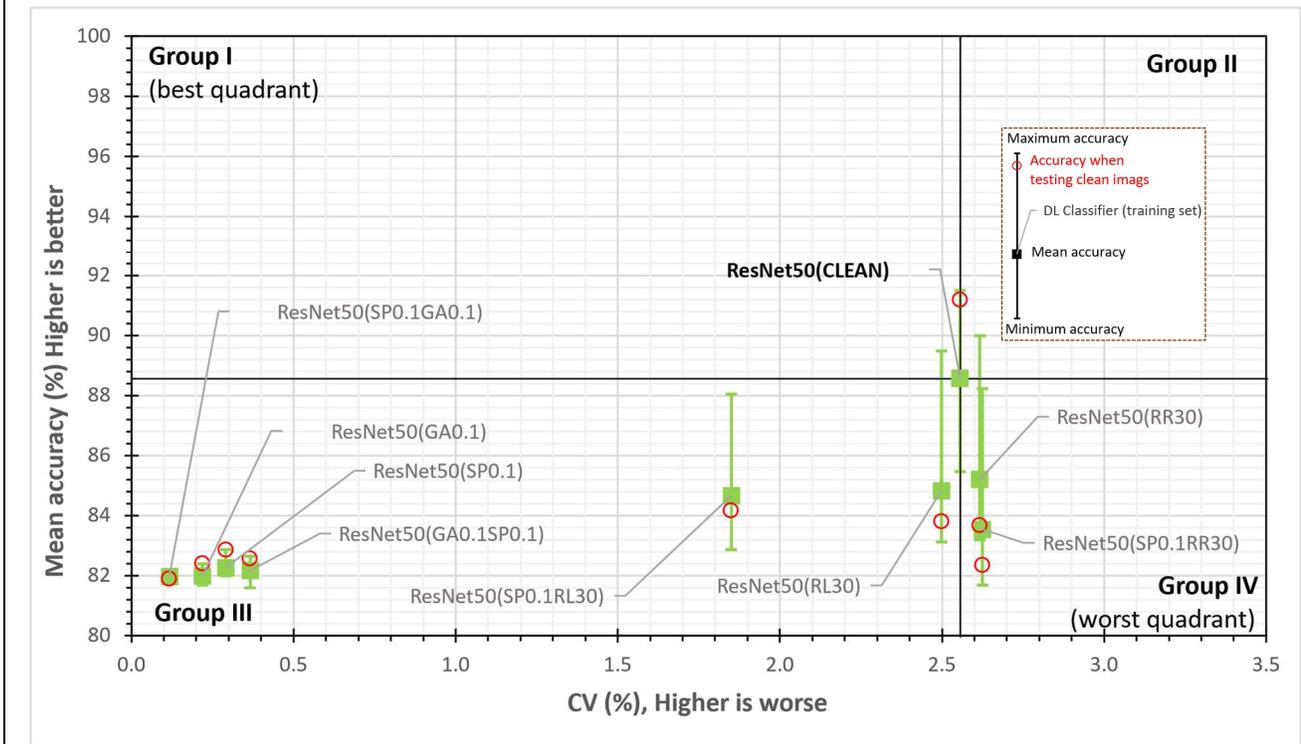

Fig. 6. The mCV plot for evaluating the robustness of ResNet50 classifiers after corrupted image groups were used to train the DL classifier on the CIFAR-10 dataset.

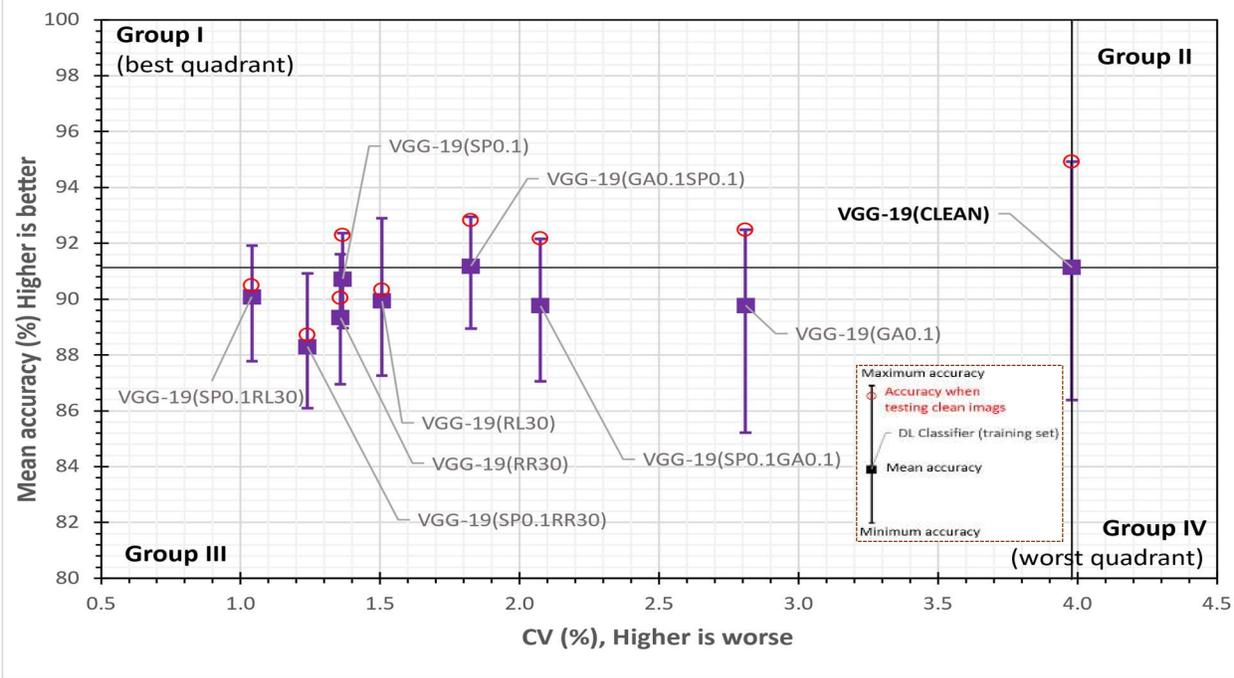

Fig. 7. The mCV plot for evaluating robustness of VGG-19 after corrupted image groups trained the DL classifier on the CIFAR-10 dataset.

TABLE III. BENCHAMRKING DEEP LEARNING CLASSIFIERS WITH PERTURBATION TYPES

| DL Classifier (training set) | CV (%) | Mean Accu(*all images*) (%) | Accu(*clean images*) (%) | Min Accu(*all images*) (%) | Max Accu(*all images*) (%) |
|---|---|---|---|---|---|
| AlexNet(clean) | 2.28 | 85.25 | 92.08 | 83.18 | 92.08 |
| AlexNet(GA0.1) | 1.50 | 89.60 | 90.9 | 87.22 | 91.7 |
| AlexNet(GA0.1SP0.1) | 1.33 | 89.75 | 90.82 | 87.62 | 91.42 |
| AlexNet(RL30) | 3.33 | 85.75 | 90.72 | 82.12 | 91.42 |
| AlexNet(RR30) | 3.07 | 86.33 | 90.74 | 82.64 | 91.7 |
| AlexNet(SP0.1) | 1.33 | 89.44 | 90.78 | 86.68 | 91.18 |
| AlexNet(SP0.1GA0.1) | 1.04 | 89.67 | 91.3 | 87.96 | 91.38 |
| AlexNet(SP0.1RL30) | 1.92 | 88.39 | 89.96 | 85.3 | 91.6 |
| AlexNet(SP0.1RR30) | 1.56 | 88.79 | 90.52 | 85.98 | 91.36 |
| ResNet50(clean) | 2.56 | 88.56 | 91.18 | 85.46 | 91.5 |
| ResNet50(GA0.1) | 0.22 | 81.98 | 82.4 | 81.68 | 82.4 |
| ResNet50(GA0.1SP0.1) | 0.37 | 82.17 | 82.56 | 81.6 | 82.64 |
| ResNet50(RL30) | 2.50 | 84.83 | 83.8 | 83.12 | 89.48 |
| ResNet50(RR30) | 2.62 | 85.20 | 83.68 | 83.2 | 90 |
| ResNet50(SP0.1) | 0.29 | 82.27 | 82.86 | 82 | 82.86 |
| ResNet50(SP0.1GA0.1) | 0.12 | 81.97 | 81.9 | 81.72 | 82.18 |
| ResNet50(SP0.1RL30) | 1.85 | 84.66 | 84.16 | 82.86 | 88.04 |
| ResNet50(SP0.1RR30) | 2.62 | 83.54 | 82.34 | 81.68 | 88.22 |
| VGG-19(clean) | 3.98 | 91.13 | 94.92 | 86.38 | 94.92 |
| VGG-19(GA0.1) | 2.81 | 89.78 | 92.48 | 85.22 | 92.48 |
| VGG-19(GA0.1SP0.1) | 1.83 | 91.18 | 92.82 | 88.94 | 92.94 |
| VGG-19(RL30) | 1.51 | 89.94 | 90.32 | 87.26 | 92.9 |
| VGG-19(RR30) | 1.36 | 89.34 | 90.04 | 86.96 | 91.6 |
| VGG-19(SP0.1) | 1.37 | 90.72 | 92.28 | 88.96 | 92.36 |
| VGG-19(SP0.1GA0.1) | 2.07 | 89.77 | 92.16 | 87.06 | 92.16 |
| VGG-19(SP0.1RL30) | 1.04 | 90.08 | 90.48 | 87.78 | 91.92 |
| VGG-19(SP0.1RR30) | 1.24 | 88.30 | 88.72 | 86.1 | 90.92 |

Note: **Accu(*clean images*)** refers to training on only clean images. **Accu(*all images*)** indicates that training includes both clean images and corrupted images. For visualizing results, see Figs. 5-7.